\documentclass[lettersize,journal]{IEEEtran}
\usepackage{cite}
\usepackage{amsmath,amssymb,amsfonts}
\usepackage{algorithmic}
\usepackage{graphicx}
\usepackage{textcomp}
\usepackage{xcolor}
\usepackage[utf8]{inputenc}
\usepackage{textgreek}
\usepackage{etoolbox}
\usepackage{hyperref} 
\usepackage{bm}
\usepackage{mathabx}
\usepackage{verbatim}
\usepackage{times}
\usepackage{subcaption}
\usepackage{nicematrix}
\usepackage{arydshln}
\usepackage{booktabs}
\usepackage{multirow}
\usepackage{tikz}
\usetikzlibrary{arrows.meta, positioning}
\usepackage{epstopdf}
\begin{document}

\title{A Consistency-Improved LiDAR-Inertial Bundle Adjustment}

\author{Xinran Li, Shuaikang Zheng$^*$, Pengcheng Zheng, Xinyang Wang, Jiacheng Li, Zhitian Li and Xudong Zou$^*$
\thanks{© 2026 IEEE.  Personal use of this material is permitted.  Permission from IEEE must be obtained for all other uses, in any current or future media, including reprinting/republishing this material for advertising or promotional purposes, creating new collective works, for resale or redistribution to servers or lists, or reuse of any copyrighted component of this work in other works.}
\thanks{This work was partly supported by Shandong Provincial Natural Science Foundation under Grant No. ZR2024ZD08. (\textit{Corresponding authors: Shuaikang Zheng; Xudong Zou})}
\thanks{Xinran Li, Xinyang Wang, Jiacheng Li and Pengcheng Zheng are with the Aerospace Information Research Institute, Chinese Academy of Sciences, Beijing 100190, China, and also with the School of Electronic, Electrical and Communication Engineering, University of Chinese Academy of Sciences, Beijing 100049, China
(email: lixinran19@mails.ucas.ac.cn)}
\thanks{Zhitian Li is with the Aerospace Information Research Institute, Chinese Academy of Sciences, Beijing 100190, China (e-mail: ztli@mail.ie.ac.cn).}
\thanks{Shuaikang Zheng are with the Aerospace Information Technology University, Jinan 250100, China.(email: zhengshuaikang18@mails.ucas.ac.cn)}
\thanks{Xudong Zou is with the Aerospace Information Research Institute, Chinese Academy of Sciences, Beijing 100190, China, also with the School of Electronic, Electrical and Communication Engineering, also with University of Chinese Academy of Sciences, Beijing 100049, China, also with the QiLu Aerospace Information Research Institute, Chinese
Academy of Sciences, Jinan 250100, China, and also with the Aerospace Information Technology University, Jinan 250100, China (e-mail: zouxd@aircas.ac.cn).}}

\markboth{}%
{Shell \MakeLowercase{\textit{et al.}}: A Sample Article Using IEEEtran.cls for IEEE Journals}


\maketitle

\begin{abstract}
Simultaneous Localization and Mapping (SLAM) using 3D LiDAR has emerged as a cornerstone for autonomous navigation in robotics. While feature-based SLAM systems have achieved impressive results by leveraging edge and planar structures, they often suffer from the inconsistent estimator associated with feature parameterization and estimated covariance. In this work, we present a consistency-improved LiDAR-inertial bundle adjustment (BA) with tailored parameterization and estimator. First, we propose a stereographic-projection representation parameterizing the planar and edge features, and conduct a comprehensive observability analysis to support its integrability with consistent estimator. Second, we implement a LiDAR-inertial BA with Maximum a Posteriori (MAP) formulation and First-Estimate Jacobians (FEJ) to preserve the accurate estimated covariance and observability properties of the system. Last, we apply our proposed BA method to a LiDAR-inertial odometry. 

\end{abstract}


\section{Introduction}
\IEEEPARstart{S}{imultaneous} Localization and Mapping (SLAM) using 3D LiDAR has become essential in autonomous systems such as unmanned ground vehicles, aerial robots, and mobile mapping platforms. High-resolution LiDAR sensors provide dense and accurate geometric information across diverse environments, enabling precise pose estimation and detailed map construction [1]–[4]. Drift over long trajectories is an inherent problem in SLAM. To reduce the cumulative drift , modern LiDAR SLAM systems commonly construct local submaps to support scan-to-map registration [2]–[6].

A prevalent approach, as demonstrated by LOAM [2] and its variants [6]–[10], represents environments via point clouds of extracted edge and planar features. However, many such systems overlook feature uncertainty induced by pose drift, resulting in suboptimal estimation. Recent efforts have addressed this by introducing bundle adjustment (BA) methods [11]–[16]. Notably, methods like BALM2 [11] and π-LSAM [12] perform bundle adjustment to jointly optimize both feature geometry and sensor poses, significantly reducing drift in mapping and localization.

Preserving estimator consistency and the correct observability properties is essential [17]; however, achieving both in LiDAR-based BA remains challenging. First, conventional parameterizations of planes and lines (e.g. Plücker coordinates [13]) can exhibit singularities and violate the observability properties of the system, especially when features pass near the origin (e.g. ground features), which may result in bad numerical stability and suboptional estimation. Second, approximated covariance and discrepant Jacobian linearization points may introduce spurious constraints (e.g. violating unobservable directions), allowing the estimator to gain spurious information and leading to inconsistency in filter-based and optimization-based systems [18]-[21].

To address these issues, we propose a LiDAR-inertial bundle adjustment reducing estimator inconsistency by preserving the accurate observability properties and
estimated covariance, and apply the BA to a real-time odometry system.

\section{Preliminaries}
\subsection{States, Propagation Model and Observation Model}

The states are composed of feature parameters and IMU states: pose, biases, velocity. In ordinate LIO system, 

\vspace{-7pt}

\begin{equation}
    \bm{x}_t = \left [ \bm{q}^T, \bm{b}_{g}^T, \bm{v}^T,\bm{b}_{a}^T, \bm{p}^T | \bm{E}_1, ...\bm{E}_m, \bm{S}_1, ...\bm{S}_n \right ]^T \label{eq1}
\end{equation} where $\bm{x}_t$ is the state vector at time t, $\bm{q}_t$  is the unit quaternion representing the orientation of the global frame in the IMU frame; $\bm{b}_{gt}$ and $\bm{b}_{at}$ are the biases of gyroscope and accelerometer measurements; $\bm{v}_t$ and $\bm{p}_t$ are velocity and position of IMU in the global frame; $\bm{E}_k$ represents the k-th edge feature parameter (e.g. Plücker coordinates $\left [ \bm{q}_{ek}, \tau_{k} \right ]$);
$\bm{S}_k$ corresponds the k-th suface feature (e.g. CP representation $\bm{\Pi}_k = OA_k  \bm{n}_k$ [13]).


Observability analysis needs minimal representation to avoid singularity, thus following [25], we rewrite \eqref{eq1} as a (15 + 4m + 3n) vector
\begin{equation}
    \bm{x}_t = \left [ \bm{s}^T, \bm{b}_{g}^T, \bm{v}^T,\bm{b}_{a}^T, \bm{p}^T | \bm{\bar{E}}_1, ...\bm{\bar{E}}_m, \bm{\bar{S}}_1, ...\bm{\bar{S}}_n \right ]^T\label{eq2m}
\end{equation} where $\bm{s}$ is the Cayley-Gibbs-Rodriguez(CGR) parameterization [32]; $\bm{\bar{S}}_k \in R^3$ and $\bm{\bar{E}}_k \in R^4$.

The propagation model of \eqref{eq2m} can be expressed as
\begin{equation}
    \frac{d}{dt}\bm{x}_t = \bm{f}_0+\bm{f}_1\bm{\omega}+\bm{f}_2\bm{a}\label{eq3}
\end{equation} where
\begin{equation}
    \begin{matrix}
    \resizebox{\columnwidth}{!}{$
        \bm{f}_0 = \begin{bmatrix}
            (-\frac{\partial \bm{s}}{\partial \bm{\theta}}\bm{b_g})^T	&
			\bm{0}^T &(\bm{g}-\bm{C}^T \bm{b}_a)^T&
			\bm{0}^T	&
			\bm{v}^T	&
			\bm{0}_{(4m+3n)\times 1}^T
        \end{bmatrix}^T
        $}
        \\
        \bm{f}_1 = \begin{bmatrix}
            \frac{\partial \bm{s}}{\partial \bm{\theta}}^T	&
			\bm{0}^T	&
			\bm{0}^T &
			\bm{0}^T	&
			\bm{0}^T	&
			\bm{0}_{(4m+3n)\times 1}^T
        \end{bmatrix}^T
        \\
        \bm{f}_2 = \begin{bmatrix}
			\bm{0}^T	&
			\bm{0}^T &
            \bm{C} &
			\bm{0}^T	&
			\bm{0}^T	&
			\bm{0}_{(4m+3n)\times 1}^T
        \end{bmatrix}^T
    \end{matrix}
\end{equation}

    where
    \begin{equation}
		\frac{\partial \bm{s}}{\partial \bm{\theta}} = \frac{1}{2} (\bm{I}+\bm{ss}^T+\left[\bm{s} \times\right])
	\end{equation} $\bm{g}$ is the gravity; $\bm{\omega}$ and $\bm{a}$ are the angular velocity and linear acceleration; $\bm{C}$ is the rotation matrix corresponding to $\bm{s}$. We apply \eqref{eq2m} to our BA despite the rotations (we use Lie group and Lie algebra to represent rotations).

    The measurements of the k-th planar feature and edge feature can be represented by
\begin{equation}
\begin{matrix}
        ^E\bm{h}_{k} = \begin{bmatrix}
        \bm{Cl}_k\\
        \bm{C}\bm{d}_k\tau_k + \bm{C}\left [ \bm{l}_k\times \right] \bm{p}
    \end{bmatrix}& ^S\bm{h}_{k} = \begin{bmatrix}
        \bm{Cn}_k\\
        \bm{p}^T\bm{n}_k+OA_k
    \end{bmatrix} 

\end{matrix}\label{hse}
\end{equation} where $\bm{l}_k$ and $\tau_k \bm{d}_k$ are direction and origin moment of the k-th edge feature, $\bm{n}_k$ and $OA_k$ are the norm and intercept of the k-th planar feature.

\subsection{Singular Feature Representation}

Taking commonly used closest-point and Plücker [13] (CPP) representations as case studies, we show that singularities degrade observability properties and numerical stability.

Edge features parameterized by Plücker coordinates may violate the observability properties. When the lines pass by the origin (i.e. $\tau_k = 0$), the unobservable directions along the global translation vanish because of the lock along the $[\bm{l}\times]\bm{d}$ . Specifically, if $rank([\bm{d}_1, ... \bm{d}_m])  = 3$, the unobservable distribution is
\begin{equation}
    \begin{matrix}
        \bm{\triangle} = span \left \{ \begin{matrix}
        \bm{n}_{R_I}, \bm{n}_{l_1} , & ... \bm{n}_{l_m}
    \end{matrix}  \right \}
    \\
    \resizebox{\columnwidth}{!}{$
        \bm{n}_{R_I} = \begin{bmatrix}
            (\frac{\partial \bm{s}}{\partial \bm{\theta}}\bm{Cg})^T
            &
            \bm{0}^T
            &
            -([\bm{v \times}]\bm{g})^T
            &
            \bm{0}^T
            &
            -([\bm{t} \times ]\bm{g})^T
            &
            -(\frac{\partial \bm{s}_{l_1}}{\partial \bm{\theta}_{l_1}}\bm{g})^T
            &
            0
            &
            \dots
            &
            -(\frac{\partial \bm{s}_{l_m}}{\partial \bm{\theta}_{l_m}}\bm{g})T
            &
            0
        \end{bmatrix}^T
        $}  
        \\
        \resizebox{\columnwidth}{!}{$
\bm{n}_{R_{l_k}} = \begin{bmatrix}
            \bm{0}_{15\times1}^T
            &
            \bm{0}^T
            &
            0
            &
            \dots
            &
            (\frac{\partial \bm{s}_{l_k}}{\partial \bm{\theta}_{l_k}}\bm{l}_k)^T
            &
            0
            &
            \dots
            &
            \bm{0}^T
            &
            0
        \end{bmatrix}^T
        $} 
    \end{matrix}\label{Pluckerin}
\end{equation} which can be verified in the same way as in the Appendix. The unobservable dimension is m + 1, corresponding to the global rotation about the direction of gravity and self-rotation about the direction of each edge feature. Global translations are erroneously observable, which can make the estimator gain information in the unobservable subspace [26].

Planar features parameterized by closest-point are prone to ill-conditioned Jacobians. Suppose that $^G\bm{p}_{scan}$ is the position of a point,
\begin{equation}
\resizebox{\columnwidth}{!}{$
       \frac{\partial (\bm{n}^T \ ^G\bm{p}_{scan}+OA)}{\partial \bm{\Pi}} = \frac{(\left \|\bm{\Pi} \right \|_2^{2}\bm{I}-\bm{\Pi}\bm{\Pi}^T) ^G\bm{p}_{scan}  }{\left \|\bm{\Pi} \right \|_2^{3}} + \frac{  \bm{\Pi} }{\left \|\bm{\Pi} \right \|_2}
       $}\label{sigCP}
\end{equation}
The elements tend to infinite when OA tends to 0. 

The above-mentioned issues are unfavorable in scenes rich in ground features, resulting in either suboptimal estimation or bad optimization. Therefore, the parameterization should avoid singularities.

\section{Consistency-Improved LiDAR-Inertial BA}
\label{proposedLIO}

In this section, we propose a stereographic-projection feature parameterization. An observability analysis
 supports its integrability with consistent estimator. Next, we develop a consistency-improved LiDAR-inertial BA with built upon MAP and FEJ, and we integrate it into the LIO back-end optimization.

\subsection{Stereographic-Projection (SP) Representation}

\begin{figure}[!t]

    \vspace{5pt}
    
    \centering
    \includegraphics[width=0.61\linewidth]{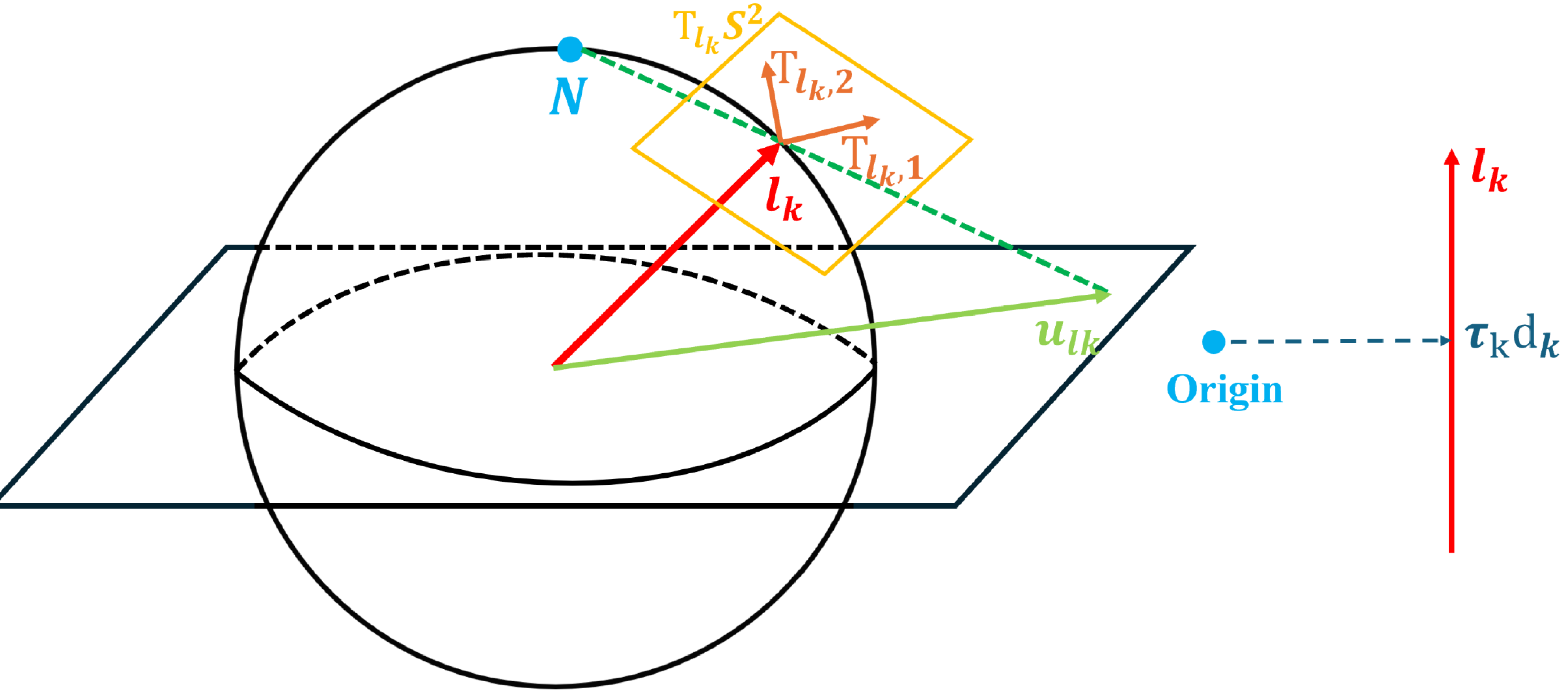}
    \caption{Stereographic Projection of $\bm{\bar{E}}_k$}
    \label{fig:enter-label1}

    \vspace{-15pt}
    
\end{figure}

Inspired by [27, Section 0.4], we apply a stereographic projection for the feature parameterization-a diffeomorphism from $R^2$ to $S^2 \setminus N$-which resolves the unit-vector parameterization. The partial derivatives of the diffeomorphism furnish tangent vectors, enabling an analytic expression for the origin moments of edge features. Following the notations defined in \eqref{hse}, denote $\bm{\bar{S}}_k = \left [ \bm{u}_{sk}, OA_{k} \right ]$ is the stereographic projection coordinate of planes, where 
\begin{equation}
\resizebox{\columnwidth}{!}{$
    \bm{n}_k = \left [ \frac{2\bm{u}_{skx}}{1+\bm{u}_{skx}^2+\bm{u}_{sky}^2}, \frac{2\bm{u}_{sky}}{1+\bm{u}_{skx}^2+\bm{u}_{sky}^2},1-\frac{2} {1+\bm{u}_{skx}^2+\bm{u}_{sky}^2} \right ]
    $}\label{planesph}
\end{equation} $\bm{\bar{E}}_k = \left [ \bm{u}_{lk}, \bm{\Lambda}_k \right ]$ such that 
\begin{equation}
    \begin{matrix}
    \resizebox{\columnwidth}{!}{$
        \bm{l}_k = \left [ \frac{2\bm{u}_{lkx}}{1+\bm{u}_{lkx}^2+\bm{u}_{lky}^2}, \frac{2\bm{u}_{lky}}{1+\bm{u}_{lkx}^2+\bm{u}_{lky}^2},1-\frac{2} {1+\bm{u}_{lkx}^2+\bm{u}_{lky}^2} \right ]^T
        $}

        \vspace{1pt}
        
        \\
        \resizebox{\columnwidth}{!}{$
        \bm{T}_{l_k, 1} = \left [ \frac{1+\bm{u}_{lky}^2 - \bm{u}_{lkx}^2}{1+\bm{u}_{lkx}^2+\bm{u}_{lky}^2}, \frac{-2\bm{u}_{lkx}\bm{u}_{lky}}{1+\bm{u}_{lkx}^2+\bm{u}_{lky}^2},\frac{2\bm{u}_{lkx}} {1+\bm{u}_{lkx}^2+\bm{u}_{lky}^2} \right ]^T = [\bm{l}_k\times]\begin{bmatrix}
            0 & 1 &\bm{u}_{lky}
        \end{bmatrix}^T \triangleq [\bm{l}_k\times]\bm{\gamma}_1
        $}

        \vspace{1pt}

        \\
                \resizebox{\columnwidth}{!}{$
        \bm{T}_{l_k, 2} = \left [ \frac{-2\bm{u}_{lkx}\bm{u}_{lky}}{1+\bm{u}_{lky}^2+\bm{u}_{lkx}^2},\frac{1+\bm{u}_{lkx}^2 - \bm{u}_{lky}^2}{1+\bm{u}_{lky}^2+\bm{u}_{lkx}^2},\frac{2\bm{u}_{lky}} {1+\bm{u}_{lkx}^2+\bm{u}_{lky}^2} \right ]^T = -[\bm{l}_k\times]\begin{bmatrix}
            1 & 0 &\bm{u}_{lkx}
        \end{bmatrix}^T \triangleq [\bm{l}_k\times]\bm{\gamma}_2
        $}

        \\
        \tau_k\bm{d}_k = \bm{\Lambda}_{kx}\bm{T}_{l_k, 1} + \bm{\Lambda}_{ky}\bm{T}_{l_k, 2} 
 \end{matrix}\label{linesph}
\end{equation} Taking \eqref{linesph} as example, the geometric intuition is given in Fig. \ref{fig:enter-label1}. We restrict $\left \| \bm{u}_{sk} \right \|_2 \le 1+\delta$, $\left \| \bm{u}_{lk} \right \|_2 \le 1+\delta$ $ (\delta>0)$ after each iteration to maintain numerical stability because the pairs $[\bm{n}_k, OA_k]$ and $[\bm{l}_k, \tau_k\bm{d}_k]$ are equivalent to $-[\bm{n}_k, OA_k]$ and $-[\bm{l}_k, \tau_k\bm{d}_k]$.
Thereby the derivatives $\left \| \bigtriangledown \bm{n}_k \right \| _{F}$,  $\left \| \bigtriangledown \bm{l}_k \right \| _{F}$ and $\left \| \bigtriangledown \bm{T}_{l_k, i} \right \| _{F}$ are all bounded with  $\sqrt{9+8\delta+4\delta^2}$. Our representation avoids any singularity, because the equivalent but not equal states $\left\{ \bm{\bar{E}}_{\alpha},\bm{\bar{E}}_{\beta} \right\}$ and $\left\{ \bm{\bar{S}}_{\alpha}, \bm{\bar{S}}_{\beta}\right\}$ can be sepreated by open balls (i.e. $B(\bm{\bar{E}}_{\alpha},\frac{1}{4})\cap B(\bm{\bar{E}}_{\beta},\frac{1}{4}) = \emptyset$, $B(\bm{\bar{S}}_{\alpha},\frac{1}{4})\cap B(\bm{\bar{S}}_{\beta},\frac{1}{4}) = \emptyset$, where $B(\bm{x},r)$ is the open ball centered at $\bm{x}$ with radius r).

Parameterizing the states through \eqref{eq2m}, closed-form unobservable distribution will be shown in the Appendix B. See proof in Appendix C.

Define r(S) = $rank([\bm{n}_1, \bm{n}_2, ... \bm{n}_n])$ and r(E) = $rank([\bm{l}_1, \bm{l}_2, ... \bm{l}_m])$. Suppose that $D$ is a declaration, the indicator function of $D$ is represented as
\begin{equation}
    1_{D} = \left\{\begin{matrix}
  1& D \ \text{is true}\\
  0&\text{otherwise}
\end{matrix}\right.\label{indicator}
\end{equation}

\begin{table}[!t]

\vspace{5pt}

\caption{\textsc{Unobservable dimension}}
\label{dimension}
\centering
\vspace{-8pt}
\begin{tabular}{lrrrrrr}
\hline
  r(S) $\setminus$ r(E)  &      0&1 & $\ge$2 \\
\hline
0 & $\setminus$& 5 + $1_{D_2}$&4 \\
1& 7 + $1_{D_1}$&4 + $1_{D_3}(1+ 1_{D_2})$ &4 \\
2&5 + $1_{D_1}$ &4 + $1_{D_3}(1+ 1_{D_2})$ &4 \\
3& 4&4 &4 \\
\hline
\end{tabular}
\vspace{-15pt}
\end{table}

Define $D_1$ is $\bm{g}\in span\left \{ \bm{n}_1, ... \bm{n}_n \right \}$; $D_2$ is $\bm{l}_1^T\bm{g} = 0$; $D_3$ is $\bm{l}_1^T\bm{n}_k = 0 (\forall 1 \le k \le n)$. The unobservable dimensions are shown in Table~\ref{dimension}.

\subsection{MAP-Based Joint BA}
\label{Joint Opt}

The keyframe IMU states and parameters of features take part in bundle adjustment. 

In our graph-based system, direct distances from local points $^I\bm{p}_{scan}$ to the planar and edge features are optimized:
\begin{equation}
\begin{matrix}
    r_{S_k} = \bm{n}_k^T (\bm{C}^T \ ^I\bm{p}_{scan} + \bm{p}) +OA_k
    \\
     \bm{d}_{E_k}= \tau_k\bm{d}_k + \left[ \bm{l}_k \times\right](\bm{C}^T \ ^I\bm{p}_{scan} + \bm{p})
    \\
    \bm{r}_{E_k} = \begin{bmatrix}
        \bm{d}_k & \bm{l}_k \times \bm{d}_k
    \end{bmatrix}^T \bm{d}_{E_k}
\end{matrix}\label{residuals}
\end{equation}

We adopt the same assumption as [11] that the additive noise of LiDAR points in the IMU frame follow a Gaussian distribution $N(0, \sigma \bm{I})$, the covariances of feature residuals in \eqref{residuals} are
\begin{equation}
    \begin{matrix}
        E(r_{S_k}^2) = \bm{n}_k^T \bm{C}^T \sigma \bm{C}\bm{n}_k = \sigma
        \\
        E(\bm{r}_{E_k}\bm{r}_{E_k}^T) =
        \\
        \begin{bmatrix} 
        \bm{d}_k & \bm{l}_k \times \bm{d}_k
    \end{bmatrix}^T\left[ \bm{l}_k \times\right]\bm{C}^T\sigma\bm{C}\left[ \bm{l}_k \times\right]\begin{bmatrix}
        \bm{d}_k & \bm{l}_k \times \bm{d}_k
    \end{bmatrix} = \sigma\bm{I}_{2\times2}
    \end{matrix}
\end{equation}

The MAP of the LiDAR feature residuals are
\vspace{-3pt}
\begin{equation}
\begin{matrix}
        p(r_{S_k}\ |\ \bm{S}_k, \bm{p}, \bm{q}) = \frac{1}{\sqrt{2\pi\sigma}}e^{-\frac{r_{S_k}^2}{2\sigma}}
        \\
        p(\bm{r}_{E_k}\ |\ \bm{E}_k, \bm{p}, \bm{q}) = \frac{1}{2\pi\sigma}e^{-\frac{\bm{r}_{E_k}^T\bm{r}_{E_k}}{2\sigma}} = \frac{1}{2\pi\sigma}e^{-\frac{\bm{d}_{E_k}^T\bm{d}_{E_k}}{2\sigma}}
\end{matrix}\label{MAP}\end{equation}

The optimal states in sliding window from N to N + W are obtained by minimizing
\begin{equation}
    \left \| \bm{r}_{P_{t_N}} \right \|_{\bm{\Sigma}_P}^2 + \sum_{n = N}^{n = N + W - 1}\left \| \bm{r}_{I_{t_n}} \right \|_{\bm{\Sigma_{I}}}^2 +  \sum_{n = N}^{n = N + W}\left \| \bm{r}_{L_{t_n}} \right \|_{\sigma\bm{I}}^2\label{residual all}
\end{equation}where $\bm{r}_{P_{t_N}}$ is the prior items from marginalization with FEJ; $\bm{r}_{I_{t_n}}$ is  the residual of the IMU constraints [5]; $\bm{r}_{L_{t_n}}$ is the residual of LiDAR constraints int the form of \eqref{residuals}.

To preserve the consistency of the system, the FEJ meathod proposed in [21] is employed. Suppose that $\bm{\delta}$ is an IMU state or feature parameter in the sliding window, $\bm{\delta}_0$ is the linearization point of the Jacobian $\bm{J}_{\delta}$, i.e. $ \bm{J}_{\delta} = \frac{\partial \bm{r}}{\partial \bm{\delta}}|_{\bm{\delta}_0}$, where $\bm{r}$ is introduced in \eqref{residual all}. $\bm{\delta}_0$ is updated if it is not the marginalization term. Otherwise $\bm{\delta}_0$ remains the same during all subsequent optimization and marginalization steps. When variables are removed by marginalization using the Schur complement, 
\begin{equation}
    \bm{H} = \bm{J}^T\bm{WJ} ,  \ \ \ \bm{b} = -\bm{J}^T\bm{Wr}
\end{equation} all set the linearization point at the initial values of the states moved into the marginalization term.

The proposed BA optimization offers the following advantages:

1). As shown in \eqref{MAP}, the covariance of the LiDAR constraints in \eqref{residual all} does not account for higher-order loss.

2). Stereographic projection parameters enables the FEJ to correctly preserve estimator consistency and improves the accuracy of the prior constraint covariance.

3). The Frobenius norms of the partial derivatives in \eqref{planesph} and \eqref{linesph} are bounded, which has numerical stability in Levenberg-Marquardt algorithm.

\subsection{LIO System Integrating Our BA}
\begin{figure}[!t]

    \vspace{5pt}
    
    \centering
    \includegraphics[width=0.5\linewidth]{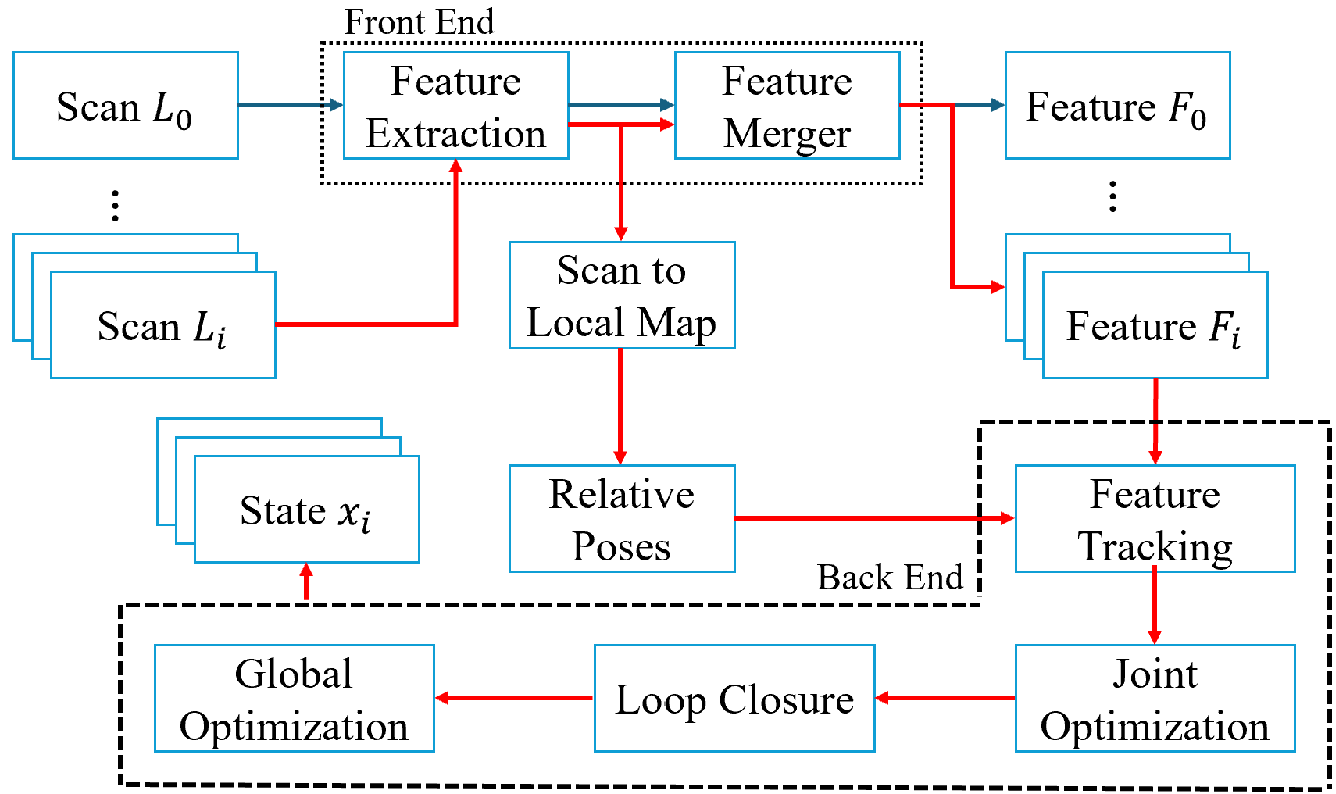}
    \caption{System Overview}
    \label{fig:enter-label2}

    \vspace{-20pt}
    
\end{figure}

\begin{figure}[!t]
\vspace{5pt}
    \centering
    \includegraphics[width=0.5\linewidth]{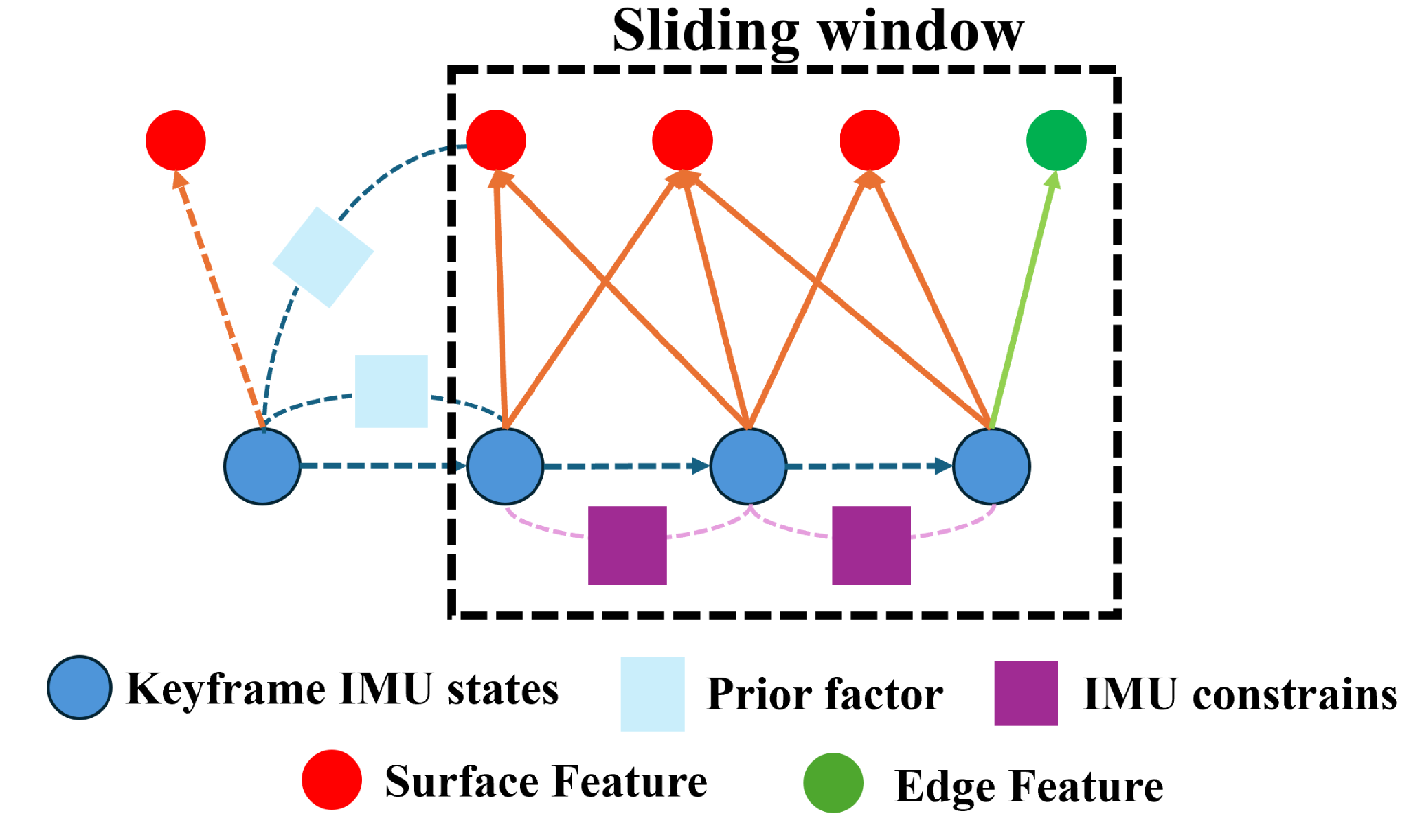}
    \caption{Joint Optimization}
    \label{fig:enter-label3}
    \vspace{-20pt}
\end{figure}

A system overview (Fig. \ref{fig:enter-label2}) and optimization strategy (Fig. \ref{fig:enter-label3}) of our LIO are shown. 


\subsubsection{Front End}

Candidate feature points, representing portions of planes and edges, are extracted from raw LiDAR scans. A local feature map is maintained using recent frames, enabling reliable relative pose estimation via scan-to-map registration. Then coplanar and collinear feature candidates are further processed to merge fragmented segments into coherent planar or linear features.

\subsubsection{Back End}

The back end operates at a lower frequency in parallel with the front end. Relative poses from scan-to-map alignment assist feature association and provide more accurate initialization than pure IMU integration. Scans containing new features may be selected as keyframes and inserted into a sliding-window joint optimization. To ensure real-time performance, keyframes are selected at most once every three frames.

Loop closure is detected by identifying nearby historical frames. When a potential loop is found, ICP is applied between the current scan and the historical local map to confirm closure. Upon confirmation, global pose graph optimization is performed.

\section*{APPENDIX A}

\vspace{-5pt}

\section*{Preliminaries}
 \vspace{-5pt}
If we have a set of smooth vector fields $\left \{\bm{f}_i\right\}_{i \in I}$ and a $C^{\infty}$ function $h$ on a manifold $X$ with local coordinate $x_1, x_2, ... ,x_N$. 

We compute the Lie derivatives
\begin{equation}
\begin{matrix}
    \mathcal{L}^0 h = h &  \mathcal{L}^{k+1}_{\bm{f}_{i_1}..\bm{f}_{i_{k+1}}}h = \bigtriangledown \mathcal{L}^{k}_{\bm{f}_{i_1}..\bm{f}_{i_{k}}}h \  \bm{f}_{i_{k+1}}
    \\
    & \bigtriangledown \lambda = [\frac{\partial \lambda}{\partial x_1}, ... \frac{\partial \lambda}{\partial x_N}]
\end{matrix}
\end{equation} for any $k\in N$ and $C^1$ function $\lambda$.

Define a standard basis of $R^3$ as $\bm{e}_1, \bm{e}_2, \bm{e}_3$.

According to local observability theorem in [31], unobservable directions span the annihilator of observability codistribution
\begin{equation}
        \mathcal{O} = \left \{ \omega \in T^*X \ | \ \omega = \bigtriangledown \mathcal{L}^{k}_{\bm{f}_{i_1}..\bm{f}_{i_{k}}}h, 0 \le j \le k, i_j \in I  \right\}
\end{equation}

In LIO system with plane and edge feature observation model, denote the unobservable direction as $\bm{n} = [\bm{n}_s^T, \bm{n}_{bg}^T, \bm{n}_v^T, \bm{n}_{ba}^T, \bm{n}_p^T | \bm{n}_{E_1}^T, ... \bm{n}_{E_m}^T, \bm{n}_{S_1}^T, ... \bm{n}_{S_n}^T]^T$ such that $\bm{n} \in \mathcal{O}^{\perp}$. 

In Appendix B, we will give the unobservable distribution $\triangle$. In Appendix C, our proof is divided into 3 steps: 

1) Prove that $\bm{n}_{bg} = \bm{n}_{ba} = \bm{0}$. 

2) The unobservable distribution $\triangle \subseteq \mathcal{O}^{\perp}$ by calculating the bases introduced in [25].

3)  $\mathcal{O}^{\perp} \subseteq \triangle$ by selecting finite Lie derivatives.

\vspace{-5pt}

\section*{APPENDIX B}

\vspace{-5pt}

\section*{Unobservable Distribution}
\vspace{-5pt}

The declaration $D_k$ is defined as:
 $D_1$ is $\bm{g}\in span\left \{ \bm{n}_1, ... \bm{n}_n \right \}$; $D_2$ is $\bm{l}_1^T\bm{g} = 0$; $D_3$ is $\bm{l}_1^T\bm{n}_k = 0 (\forall 1 \le k \le n)$. Define r(S) = $rank([\bm{n}_1, \bm{n}_2, ... \bm{n}_n])$ and r(E) = $rank([\bm{l}_1, \bm{l}_2, ... \bm{l}_m])$. Suppose that $D$ is a declaration, the indicator function of $D$ is represented as
\begin{equation}
    1_{D} = \left\{\begin{matrix}
  1& D \ \text{is true}\\
  0&\text{otherwise}
\end{matrix}\right.\label{indicator}
\end{equation}.

Define the derivatives of stereographic projection coordinate $[x(u_x,u_y),y(u_x,u_y),z(u_x,u_y)]\in S^2 \setminus  N$  as a $3\times2$ matrix
\begin{equation}
\resizebox{\columnwidth}{!}{$
    \bm{H}_{[x,y,z]} =  \frac{\partial [x , y, z]^T}{\partial[u_x,u_y]}= 2\begin{bmatrix}
        \frac{1+u_y^2-u_x^2}{(1+u_y^2+u_x^2)^2} &\frac{-2u_xu_y}{(1+u_y^2+u_x^2)^2}& \frac{2u_x}{(1+u_y^2+u_x^2)^2} \\ \frac{-2u_xu_y}{(1+u_y^2+u_x^2)^2} 
          & \frac{1+u_x^2-u_y^2}{(1+u_y^2+u_x^2)^2}
 & \frac{2u_y}{(1+u_y^2+u_x^2)^2}
    \end{bmatrix}^T
    $}
\end{equation} it is trivial to check $rank(\bm{H}_{[x,y,z]}) = 2$ and left null space of $\bm{H}_{[x,y,z]}$ is [x, y ,z].

Denote the unobservable distribution as
\begin{equation}
   \triangle =  col \left \{ \begin{bmatrix}
        \bm{n}_1 & \bm{n}_2 & \dots &\bm{n}_u
    \end{bmatrix} \right \} \label{eq18}
\end{equation} where $u = dim(\triangle)$. $\forall 0\le v \le u$, $\bm{n}_v$ has the form
\begin{equation}
    \begin{bmatrix}
        \bm{n}_{I, v}^T &\bm{n}_{E_1,v}^T & \dots & \bm{n}_{E_m,v}^T &  \bm{n}_{S_1,v}^T&\dots &\bm{n}_{S_n,v}^T
    \end{bmatrix}^T
\end{equation} and denote $\bm{H}_{l_j}^{+} = (\bm{H}_{l_j}^T\bm{H}_{l_j})^{-1}\bm{H}_{l_j}^T$
\begin{equation}
\begin{matrix}
        \begin{bmatrix}
        \bm{n}_{I, 1}, & \bm{n}_{I,2}, & \bm{n}_{I,3}
    \end{bmatrix} = [\bm{0}^T \ \bm{0}^T \ \bm{0}^T \ \bm{0}^T \ \bm{I}_{3\times3} ]^T
    \\
    \begin{bmatrix}
        \bm{n}_{S_j, 1}, & \bm{n}_{S_j,2}, & \bm{n}_{S_j,3} 
    \end{bmatrix}=  [\bm{0}, \ -\bm{n}_j]^T
    \\
    \resizebox{\columnwidth}{!}{$
\begin{bmatrix}
        \bm{n}_{E_j, 1}, & \bm{n}_{E_j,2}, & \bm{n}_{E_j,3} 
    \end{bmatrix} = [\bm{0}, -(1+\bm{u}_{ljx}^2+\bm{u}_{ljy}^2)(\bm{H}_{l_j}^{+}[\bm{l}_j\times])^T ]^T
    $}
\end{matrix}\label{translation}
\end{equation}

\subsubsection{r(E) = 0 , r(S) = 1}

 When $1_{D_1} = 0$, $u = 7$
\begin{equation}
\begin{matrix}
        \bm{n}_{I, 4} = [\bm{0}^T \ \bm{0}^T \ (\bm{n}_1^{\perp})_1^T \ \bm{0}^T \ \bm{0}^T ]^T
        \\
        \bm{n}_{I, 5} = [\bm{0}^T \ \bm{0}^T \ (\bm{n}_2^{\perp})_1^T \ \bm{0}^T \ \bm{0}^T ]^T
        \\
        \bm{n}_{I, 6} = [(\frac{\partial \bm{s}}{\partial \bm{\theta}}\bm{Cn}_1)^T \ \bm{0}^T \ \bm{0}^T \ \bm{0}^T \ \bm{0}^T]^T
        \\
        \bm{n}_{I, 7} = [(\frac{\partial \bm{s}}{\partial \bm{\theta}}\bm{Cg})^T \ \bm{0}^T \ \bm{v}^T[\bm{n}_1\times]\bm{g}\bm{n_1}^T \ \bm{0}^T \ \bm{0}^T]^T
\end{matrix} \label{rs=1}
\end{equation} where $(\bm{n}_1^{\perp})_1$ and $(\bm{n}_1^{\perp})_2$ are two orthogonal unit vectors such that 
$(\bm{n}_1^{\perp})_i^T\bm{n}_1 = 0$.

\vspace{5pt}

\begin{equation}
    \begin{matrix}
        \bm{n}_{S_j, 4} = \bm{n}_{S_j, 5} = \bm{n}_{S_j, 6} = \bm{0}
        \\
        \bm{n}_{S_j, 7} = [ -(\bm{H}_{n_j}^+[\bm{n}_j\times]\bm{g})^T, \bm{p}^T[\bm{n}_j\times]\bm{g} ]^T
    \end{matrix}\label{rs=1'}
\end{equation} $\forall 1\le j \le n$, where $\bm{H}_{n_j}^+=(\bm{H}_{n_j}^T\bm{H}_{n_j})^{-1}\bm{H}_{n_j}^T$.

When $1_{D_1} = 1$, $u = 8$,
\begin{equation}
\begin{matrix}
        [\bm{n}_{I,6}, \ \bm{n}_{I,7}, \ \bm{n}_{I,8}] = [(\frac{\partial \bm{s}}{\partial \bm{\theta}}\bm{C})^T \ \bm{0}^T \ [\bm{v}\times] \ \bm{0}^T \ \bm{0}^T]^T
        \\
        \resizebox{\columnwidth}{!}{$
            [\bm{n}_{S_j,6}, \ \bm{n}_{S_j,7}, \ \bm{n}_{S_j,8}] = [ -(\bm{H}_{n_j}^+[\bm{n}_j\times])^T, (\bm{p}^T[\bm{n}_j\times])^T ]^T
        $}
\end{matrix}
\end{equation} where $\bm{n}_4$ and $\bm{n}_5$ are same as \eqref{rs=1} and \eqref{rs=1'}.

\subsubsection{r(E) = 0, r(S) = 2}

Suppose that
\begin{equation}
    span\left\{ \bm{n}_1, ...\bm{n}_n \right\} = span\left\{ \bm{n}_1, \bm{n}_2 \right\} \label{assumption1}
\end{equation}

When $1_{D_1} = 0$, $u = 5$
\begin{equation}
    \begin{matrix}
        
        \bm{n}_{I, 4} = [\bm{0}^T \ \bm{0}^T \ ([\bm{n}_1\times]\bm{n}_2)^T \ \bm{0}^T \ \bm{0}^T ]^T
        \\
        \bm{n}_{I, 5} = [(\frac{\partial \bm{s}}{\partial \bm{\theta}}\bm{Cg})^T \ \bm{0}^T \ \mathcal{V}\bm{N}^{-1} \ \bm{0}^T \ \bm{0}^T]^T
    \end{matrix}\label{rs=2}
\end{equation} where $\bm{N} = [\bm{n}_1, \bm{n}_2, [\bm{n}_1 \times]\bm{n}_2]$ and $\mathcal{V} = [\bm{v}^T[\bm{n}_1 \times]\bm{g}, \bm{v}^T[\bm{n}_2 \times]\bm{g}, 0]$
\begin{equation}
    \begin{matrix}      
        \bm{n}_{S_j, 4} = \bm{0} \\
        \bm{n}_{S_j, 5} = [ -(\bm{H}_{n_j}^+[\bm{n}_j\times]\bm{g})^T, \bm{p}^T[\bm{n}_j\times]\bm{g} ]^T
    \end{matrix}\label{rs=2'}
\end{equation}

 When $1_{D_1} = 1$, $u = 6$, $\bm{n}_{I,v} (v\le4)$ are same as above.
\begin{equation}
    \begin{matrix}
            \bm{n}_{I,5} = [(\frac{\partial \bm{s}}{\partial \bm{\theta}}\bm{Cn}_1)^T \ \bm{0}^T \ \mathcal{U}_1^T \ \bm{0}^T \ \bm{0}^T]^T
            \\
            \bm{n}_{I,6} = [(\frac{\partial \bm{s}}{\partial \bm{\theta}}\bm{Cn}_2)^T \ \bm{0}^T \ \mathcal{U}_2^T \ \bm{0}^T \ \bm{0}^T]^T
    \end{matrix}
\end{equation} where $\mathcal{U}_1 = \frac{\bm{v}^T[\bm{n}_2\times]\bm{n}_1}{\bm{n}_2^T[\bm{n}_1\times][\bm{n}_1\times]\bm{n}_2}[\bm{n}_1\times][\bm{n}_1\times]\bm{n}_2$ and $\mathcal{U}_2 = \frac{\bm{v}^T[\bm{n}_1\times]\bm{n}_2}{\bm{n}_1^T[\bm{n}_2\times][\bm{n}_2\times]\bm{n}_1}[\bm{n}_2\times][\bm{n}_2\times]\bm{n}_1$.
\begin{equation}
    \begin{matrix}
            \bm{n}_{S_j, 5} = [ -(\bm{H}_{n_j}^+[\bm{n}_j\times]\bm{n}_1)^T, (\bm{p}^T[\bm{n}_j\times]\bm{n}_1)^T ]^T
            \\
            \bm{n}_{S_j, 6} = [ -(\bm{H}_{n_j}^+[\bm{n}_j\times]\bm{n}_2)^T, (\bm{p}^T[\bm{n}_j\times]\bm{n}_2)^T ]^T
    \end{matrix}
\end{equation} $\bm{n}_4$ is same as \eqref{rs=2} and \eqref{rs=2'}.

\subsubsection{r(S) = 3 or r(E) $\ge$ 2}

The LIO system preserves at least 4 unobservable directions corresponding the global translations and global rotation about gravity.
\begin{equation}
    \begin{matrix}
        \bm{n}_{I, 4} = [(\frac{\partial \bm{s}}{\partial \bm{\theta}}\bm{Cg})^T \ \bm{0}^T \ -([\bm{v}\times]\bm{g})^T \ \bm{0}^T \ -([\bm{p}\times]\bm{g})^T]^T
        \\

        \resizebox{\columnwidth}{!}{$
        \bm{n}_{E_j, 4} = [ -(\bm{H}_{l_j}^{+}[\bm{l}_j\times]\bm{g})^T, \frac{1}{1+\bm{u}_{l1x}^2+\bm{u}_{l1y}^2}(\bm{H}_{l_j}^{+}[\bm{l}_j\times](\begin{bmatrix}
                0 & 0 &-\bm{\Lambda}_{jy}
                \\
                0&0&\bm{\Lambda}_{jx}
            \end{bmatrix}^T\bm{H}_{l_j}^{+}[\bm{l} \times] -[(\bm{\Lambda}_{jx}\bm{\gamma}_{j1} +\bm{\Lambda}_{jy}\bm{\gamma}_{j2})\times] )\bm{g})^T ]^T
        $} 
        \\
        \bm{n}_{S_j, 4} = [ -(\bm{H}_{n_j}^+[\bm{n}_j\times]\bm{g})^T, 0 ]^T
    \end{matrix} \label{minimum_unobservable}
\end{equation}  

\subsubsection{r(E) = 1}

In addition to the directions declared in \eqref{translation} and \eqref{minimum_unobservable}, $1_{D_3} = 1$ and $1_{D_2} 1_{D_3} = 1$ introduce 2 more unobservable directions
\begin{equation}
    \begin{matrix}
            \resizebox{\columnwidth}{!}{$
        \bm{n}_{I, 5} = [(\frac{\partial \bm{s}}{\partial \bm{\theta}}\bm{C}\mathcal{L_G})^T \ \bm{0}^T \ -([\bm{v}\times]\mathcal{L_G})^T \ \bm{0}^T \ -([\bm{p}\times]\mathcal{L_G})^T]^T
        $}  
        \\
        \bm{n}_{I, 6} = [\bm{0}^T \ \bm{0}^T \ \bm{l}_1^T \ \bm{0}^T \ \bm{0}^T ]^T
        \\

        \resizebox{\columnwidth}{!}{$
        \bm{n}_{E_j, 5} = [ -(\bm{H}_{l_j}^{+}[\bm{l}_j\times]\mathcal{L_G})^T, \frac{1}{1+\bm{u}_{l1x}^2+\bm{u}_{l1y}^2}(\bm{H}_{l_j}^{+}[\bm{l}_j\times](\begin{bmatrix}
                0 & 0 &-\bm{\Lambda}_{jy}
                \\
                0&0&\bm{\Lambda}_{jx}
            \end{bmatrix}^T\bm{H}_{l_j}^{+}[\bm{l} \times] -[(\bm{\Lambda}_{jx}\bm{\gamma}_{j1} +\bm{\Lambda}_{jy}\bm{\gamma}_{j2})\times] )\mathcal{L_G})^T ]^T
        $} 
        \\

        \bm{n}_{S_j, 5} = [ -(\bm{H}_{n_j}^+[\bm{n}_j\times]\mathcal{L_G})^T, 0 ]^T
        \\
        \bm{n}_{E_j, 6} = \bm{n}_{S_j, 6} = \bm{0}
        
    \end{matrix}
\end{equation} where $\mathcal{L_G} = [\bm{l}\times]\bm{g}$.

\vspace{-2pt}

\section*{APPENDIX C}

\vspace{-5pt}

\section*{Sketch of the Proof}

In this section, we will use Einstein summation convention for simplifying expressions of summation.

\subsubsection*{1) Subsystem with Only One Feature}

Define functions $\bm{F}$ and $\bm{G}$ from the state space of LIO system to its subspace such that $\bm{F}(\bm{x}) = \left [ \bm{s}^T, \bm{b}_{g}^T, \bm{v}^T,\bm{b}_{a}^T, \bm{p}^T | \bm{\bar{E}}_1 \right ]^T$ and $\bm{G}(\bm{x})=\left [ \bm{s}^T, \bm{b}_{g}^T, \bm{v}^T,\bm{b}_{a}^T, \bm{p}^T | \bm{S}_1 \right ]^T$. The observability codistributions of these two subsystem are $\mathcal{O}_E$ and $\mathcal{O}_S$. The sets of Lie derivatives corresponding to the plane and edge observation $^{S} \bm{h}_1$ and $^E \bm{h}_1$ can be expressed as
\begin{equation}
\begin{matrix}
        \bm{F}^{*} \mathcal{O}_E = \left\{ \bm{F}^*\bm{\omega} | \bm{\omega} \in \mathcal{O}_E \right\},  \    \bm{G}^{*} \mathcal{O}_S = \left\{ \bm{G}^*\bm{\omega} | \bm{\omega} \in \mathcal{O}_S \right\}
\end{matrix}
\end{equation}

If any vector field $\bm{n}_{sub}$ $\in$ $\mathcal{O}_E^{\perp}$ and $\mathcal{O}_S^{\perp}$ satisfies that $\bm{n}_{bg} = \bm{n}_{ba} = \bm{0}$, then  $\forall \bm{n}$ $\in$ $\mathcal{O}^{\perp}$, $\bm{n}_{bg} = \bm{n}_{ba} = \bm{0}$. It is because $\forall \bm{n} \in \mathcal{O}^{\perp}$ and $\bm{\omega} \in \mathcal{O}_E$ or $\bm{\mu} \in \mathcal{O}_S$, $\bm{\omega}(\bm{F}_{*}\bm{n}) = \bm{\mu}(\bm{G}_{*}\bm{n}) = \bm{F}^{*}\bm{\omega}(\bm{n}) = \bm{G}^{*}\bm{\mu}(\bm{n}) = 0$


Now we will focus on the observability of the subsystem.

\subsubsection*{\textbf{Lemma C} } If $[\bm{x}\times][\bm{y}\times] = \bm{0}$ and $\bm{y} \ne 0$, then $\bm{x} =  \bm{0}$.

proof. If $\bm{x} \ne \bm{0}, \ [\bm{x}\times][\bm{x}\times]\bm{y} =-[\bm{x}\times][\bm{y}\times]\bm{x} = \bm{0}$. The solution is that $\bm{x} = k\bm{y} (k \ne 0)$. Then $k[\bm{y}\times][\bm{y}\times] = \bm{0 }$, which is contrast to $k\ne0$.
\hfill $\blacksquare$

\subsubsection*{\textbf{(1.1)} Plane} 
        
        


Define $\bm{\lambda}_{S_1}^k = \mathcal{L}^2_{\bm{f}_0\bm{f}_1^k}\bm{h} + \bm{b}_g^i\mathcal{L}^2_{\bm{f}_1^i\bm{f}_1^k}\bm{h} = \bm{0}$
\begin{equation}
\begin{matrix}
\resizebox{\columnwidth}{!}{$
    \bm{0} = \bigtriangledown \bm{\lambda}_{S_1}^k
    = \bigtriangledown\mathcal{L}^2_{\bm{f}_0\bm{f}_1^k}\bm{h}+\bm{b}_g^i\bigtriangledown\mathcal{L}^2_{\bm{f}_1^i\bm{f}_1^k}\bm{h}+\mathcal{L}^2_{\bm{f}_1^i\bm{f}_1^k}\bm{h}\bigtriangledown\bm{b}_g^i
    $}
\end{matrix} 
\end{equation}

Therefore, $\forall \bm{n} \in \mathcal{O}^{\perp}$
\begin{equation}
    (\mathcal{L}^2_{\bm{f}_1^i\bm{f}_1^k}\bm{h}\bigtriangledown\bm{b}_g^i) \  \ \bm{n} = \bm{0}
\end{equation} then $[\bm{n}_{bg}\times][\bm{e}_k\times]\bm{Cn}_1 = \bm{0 }(\forall k = 1, 2, 3)$. We deduce that $[\bm{n}_{bg}\times][\bm{Cn}_1\times] = \bm{0 }$. By \textbf{Lemma C}, $\bm{n}_{ba} = \bm{0}$.

Denote $\bm{\lambda}_{S_0} = \mathcal{L}^2_{\bm{f}_0\bm{f}_0}\bm{h} + \bm{b}_g^i\mathcal{L}^2_{\bm{f}_1^i\bm{f}_0}\bm{h} = \begin{bmatrix}
    \bm{0}\\
    \bm{n}_1^T(\bm{g}-\bm{C}^T\bm{b}_{a})
\end{bmatrix}$ and $\bm{\lambda}_{S_2}^k = \mathcal{L}^2_{\bm{f}_0\bm{f}_2^k}\bm{h} + \bm{b}_g^i\mathcal{L}^2_{\bm{f}_1^i\bm{f}_2^k}\bm{h} = \begin{bmatrix}
    \bm{0}\\
    \bm{n}_1^T\bm{C}^T\bm{e}_{k}
\end{bmatrix}$. Then
\begin{equation}
        \resizebox{\columnwidth}{!}{$
            \bigtriangledown \bm{\lambda}_{S_0} + \bm{b}_a^k\bigtriangledown\bm{\lambda}_{S_2}^k =  
        \begin{bmatrix}
            \bm{0} & \bm{0}& \bm{0} & \bm{0} & \bm{0} & \bm{0} & \bm{0} \\
            \bm{0} & -\bm{n}_1^T\bm{C}^T &\bm{0}  & \bm{0} & \bm{0} & \bm{g}^T\bm{H}_{n_1} & 0
        \end{bmatrix}
        $}\label{eq38}
\end{equation}

Therefore $\mathcal{L}^1_{\bm{f}_1^l}\bm{\lambda}_{S_0} + \bm{b}_a^k \mathcal{L}^1_{\bm{f}_1^l} \bm{\lambda}_{S_2}^k = \bm{0}$ and 
\begin{equation}
    \mathcal{L}^2_{\bm{f}_1^l\bm{f}_1^m}\bm{\lambda}_{S_0} + \bm{b}_a^k \mathcal{L}^2_{\bm{f}_1^l\bm{f}_1^m} \bm{\lambda}_{S_2}^k = \bm{0} \label{ba0}
\end{equation}

\eqref{ba0} can be rewrite as
\begin{equation}
\resizebox{\columnwidth}{!}{$
        \mathcal{L}^4_{\bm{f}_0\bm{f}_0\bm{f}_1^l\bm{f}_1^m}\bm{h} + \bm{b}_g^i \mathcal{L}^4_{\bm{f}_1^i\bm{f}_0\bm{f}_1^l\bm{f}_1^m}\bm{h} + \bm{b}_a^k \mathcal{L}^4_{\bm{f}_0\bm{f}_2^k\bm{f}_1^l\bm{f}_1^m}\bm{h}+ \bm{b}_a^k \bm{b}_g^i\mathcal{L}^4_{\bm{f}_1^i\bm{f}_2^k\bm{f}_1^l\bm{f}_1^m}\bm{h} = \bm{0}
        $} \label{nba}
\end{equation} where $\mathcal{L}^4_{\bm{f}_0\bm{f}_2^k\bm{f}_1^l\bm{f}_1^m}\bm{h} = \begin{bmatrix}
    \bm{0}
    \\
    -\bm{e}_k^T[\bm{e}_l\times][\bm{Cn}_1\times]\bm{e}_m
\end{bmatrix}$.

Take the derivative of \eqref{nba} and multiply $\bm{n}$, \begin{equation}
-\bm{e}_k^T[\bm{e}_l\times][\bm{Cn}_1\times]\bm{e}_m\bigtriangledown\bm{b}_{a}^k \ \ \bm{n} = \bm{0}
\end{equation} it can be concluded that $\bm{n}_{ba}^T[\bm{e_l}\times][\bm{Cn}_1\times]\bm{e}_m = 0$ $(\forall l, m \in \left\{ 1,2,3\right\})$. By \textbf{Lemma C}, $\bm{n}_{ba} = \bm{0}$.

\subsubsection*{\textbf{(1.2)} Edge}

Define $\bm{d}_{E_k} = \bm{C}\left[\bm{l}_k \times \right](\bm{\Lambda}_{kx}\bm{\gamma}_{k1} +\bm{\Lambda}_{ky}\bm{\gamma}_{k2}+ \bm{p})$, $\bm{\lambda}_{E_1}^k = \mathcal{L}^2_{\bm{f}_0\bm{f}_1^k}\bm{h} + \bm{b}_g^i\mathcal{L}^2_{\bm{f}_1^i\bm{f}_1^k}\bm{h} = \begin{bmatrix}
    \bm{0}
    \\
    -[\bm{e}_k \times]\bm{C}\left[\bm{l}_1 \times \right]\bm{v}
\end{bmatrix}$
\begin{equation}
    \begin{matrix}
            \resizebox{\columnwidth}{!}{$
    \begin{bmatrix}
        \bm{0}
        \\
        *
    \end{bmatrix} = \bigtriangledown \bm{\lambda}_{E_1}^k
    = \bigtriangledown\mathcal{L}^2_{\bm{f}_0\bm{f}_1^k}\bm{h}+\bm{b}_g^i\bigtriangledown\mathcal{L}^2_{\bm{f}_1^i\bm{f}_1^k}\bm{h}+\mathcal{L}^2_{\bm{f}_1^i\bm{f}_1^k}\bm{h}\bigtriangledown\bm{b}_g^i
    $}
\end{matrix} 
\end{equation} then $[\bm{n}_{bg}\times][\bm{e}_k\times]\bm{Cl}_1 = \bm{0 }(\forall k = 1, 2, 3)$. We deduce that $[\bm{n}_{bg}\times][\bm{Cl}_1\times] = \bm{0 }$. By \textbf{Lemma C}, $\bm{n}_{bg} = \bm{0}$.
\begin{equation}
    \begin{matrix}
    \resizebox{\columnwidth}{!}{$
        \bm{\lambda}_{E_0} = \mathcal{L}^2_{\bm{f}_0\bm{f}_0}\bm{h} + \bm{b}_g^i\mathcal{L}^2_{\bm{f}_1^i\bm{f}_0} = \begin{bmatrix}
    \bm{0}\\
    [\bm{b}_g \times]\bm{C}\left[\bm{l}_1 \times \right]\bm{v}+\bm{C}[\bm{l}_1\times]\bm{g} + [\bm{b}_a\times]\bm{Cl}_1
\end{bmatrix}
$}
\\
\bm{\lambda}_{E_2}^k = \mathcal{L}^2_{\bm{f}_0\bm{f}_2^k}\bm{h} + \bm{b}_g^i\mathcal{L}^2_{\bm{f}_1^i\bm{f}_2^k} = \begin{bmatrix}
    \bm{0}\\
    -[\bm{e}_{k}\times]\bm{C}\bm{l}_1
\end{bmatrix}
    \end{matrix}
\end{equation}
\begin{equation}
\begin{matrix}
        \bigtriangledown \bm{\lambda}_{E_0} + \bm{b}_{a}^k\bigtriangledown \bm{\lambda}_{E_2}^k + \bm{b}_{g}^l\bigtriangledown \bm{\lambda}_{E_1}^l
        \\
        \resizebox{\columnwidth}{!}{$
            =\begin{bmatrix}
            \bm{0} & \bm{0}& \bm{0} & \bm{0} & \bm{0} & \bm{0} & \bm{0} \\
            [\bm{C}\left[\bm{l}_1 \times \right]\bm{g} \times]\frac{\partial\bm{\theta}}{\partial \bm{s}} & -[\bm{C}\left[\bm{l}_1 \times \right]\bm{v} \times] & \bm{0}& -[\bm{C}\bm{l}_1 \times] & \bm{0} & -\bm{C}[\bm{g}\times]\bm{H}_{l_1} & 0
        \end{bmatrix}
        $}
\end{matrix}\label{eq44}
\end{equation}multiplying $\bm{n}$
\begin{equation}
    \bm{0} = -[\bm{Cl}_1\times]\bm{n}_{ba} + \bigtriangledown (\bm{C}[\bm{l}_1\times]\bm{g}) \bm{n} \label{eq45}
\end{equation} multiplying $\bm{f}_{1}^h$, $\mathcal{L}^1_{\bm{f}_1^h}\bm{\lambda}_{E_0}+\bm{b}_{a}^k \mathcal{L}^1_{\bm{f}_1^h}\bm{\lambda}_{E_2}^k+\bm{b}_{g}^l\mathcal{L}^1_{\bm{f}_1^h}\bm{\lambda}_{E_1}^l = \begin{bmatrix}
    \bm{0}\\  -[\bm{e}_h\times]\bm{C}[\bm{l}_1\times]\bm{g}
\end{bmatrix}$. Taking derivatives and multiplying $\bm{n}$, 
\begin{equation}
    [\bm{n}_{ba}\times][\bm{e}_{h}\times]\bm{Cl}_1 = -[\bm{e}_{h}\times]\bigtriangledown (\bm{C}[\bm{l}_1\times]\bm{g})\bm{n}\label{eq46}
\end{equation}

Combining \eqref{eq45}-\eqref{eq46} by Jacobi identity, $[\bm{Cl}_1\times][\bm{n}_{ba}\times]\bm{e}_{h}=\bm{0} \ (\forall h)$. By \textbf{Lemma C} that $\bm{n}_{ba} = \bm{0}$.

\subsubsection*{2) Basis Function Calculation}

The basis functions are variables in the observability codistribution. Besides $\bigtriangledown \bm{b}_{g}$ and $\bigtriangledown \bm{b}_a$, define a finite generated codistribution
\setcounter{MaxMatrixCols}{20} 
\begin{equation}
\resizebox{\columnwidth}{!}{$
\begin{matrix}
    \begin{bmatrix}
    \bar{\mathcal{O}}_{S_k}
    \\
    \bar{\mathcal{O}}_{E_k}
\end{bmatrix} = 
\begin{bmatrix}
    \bigtriangledown (\bm{Cn}_k) \\ \bigtriangledown (\bm{p}^T\bm{n}_k+OA_k) \\ \bigtriangledown (\bm{n}_k^T\bm{v}) \\\bigtriangledown (\bm{n}_k^T\bm{g}) \\ \bigtriangledown (\bm{Cl}_k) \\
    \bigtriangledown \bm{d}_{E_k} \\\bigtriangledown (\bm{C}[\bm{l}_{k}\times]\bm{v}) \\ \bigtriangledown (\bm{C}[\bm{l}_{k}\times]\bm{g})
\end{bmatrix}  = \begin{bmatrix}
    [\bm{Cn}_k\times]\frac{\partial\bm{\theta}}{\partial\bm{s}} & \bm{0} &\bm{0} &\bm{0} &\bm{0} | \dots & \bm{0} & \bm{0} & \dots &\bm{CH}_{n_k} &0 &\dots
    \\
    \bm{0} & \bm{0} &\bm{0} &\bm{0} &\bm{n}_k^T | \dots& \bm{0} & \bm{0} & \dots &\bm{p}^T\bm{H}_{n_k} &1 &\dots 
    \\
    \bm{0} & \bm{0} & \bm{n}_k^T &\bm{0} &\bm{0} | \dots & \bm{0} & \bm{0}& \dots&\bm{v}^T\bm{H}_{n_k} &0 &\dots
    \\
    \bm{0} & \bm{0} & \bm{0} &\bm{0} &\bm{0} | \dots & \bm{0} & \bm{0}& \dots&\bm{g}^T\bm{H}_{n_k} &0 &\dots 
    \\
    [\bm{Cl}_k \times]\frac{\partial\bm{\theta}}{\partial \bm{s}} & \bm{0} & \bm{0} & \bm{0} & \bm{0}|\dots & \bm{CH}_{l_k} & \bm{0}&\dots& \bm{0} & 0&\dots
    \\
            
    [\bm{d}_{E_k} \times]\frac{\partial\bm{\theta}}{\partial \bm{s}} & \bm{0} & \bm{0} & \bm{0} & \bm{C}\left[\bm{l}_k \times \right] | \dots &-\bm{C}[(\bm{\Lambda}_{kx}\bm{\gamma}_{k1} +\bm{\Lambda}_{ky}\bm{\gamma}_{k2}+ \bm{p}) \times]\bm{H}_{l_k}+\bm{C}[\bm{l}_k\times]\begin{bmatrix}
                0 & 0 &-\bm{\Lambda}_{ky}
                \\
                0&0&\bm{\Lambda}_{kx}
            \end{bmatrix}^T&(1+\bm{u}_{lkx}^2+\bm{u}_{lky}^2)\bm{C}\bm{H}_{l_k}& \dots &\bm{0} & 0&\dots
    \\
    [\bm{C}[\bm{l}_{k}\times]\bm{v} \times]\frac{\partial\bm{\theta}}{\partial \bm{s}} & \bm{0} & \bm{C}\left[\bm{l}_k \times \right] & \bm{0} & \bm{0} | \dots & -\bm{C}[\bm{v} \times]\bm{H}_{l_k} & \bm{0}&\dots &\bm{0}&0&\dots
    \\
    [\bm{C}[\bm{l}_{k}\times]\bm{g} \times]\frac{\partial\bm{\theta}}{\partial \bm{s}} & \bm{0} & \bm{0} & \bm{0} & \bm{0} | \dots & -\bm{C}[\bm{g} \times]\bm{H}_{l_k} & \bm{0}&\dots &\bm{0}&0&\dots    
\end{bmatrix}  
\end{matrix}
        $}\label{basis}
\end{equation}It is trivial to verify the unobservable vectors annihilating \eqref{basis}. Because all 1-form in $\mathcal{O}$ is the linear combination of the basis functions, $\triangle \subseteq \mathcal{O}^{\perp}$.

\subsubsection*{3) Rank Calculation}

Basis functions in \eqref{basis} can be obtained by finite linear combination of the Lie derivatives $\bigtriangledown \bm{b}_g, \   \bigtriangledown \bm{b}_a$ and $ \bigtriangledown\mathcal{L}^l_{f_{j_1}...f_{j_l}} \ ^E \bm{h}_k, \bigtriangledown\mathcal{L}^l_{f_{i_1}...f_{i_l}} \ ^S \bm{h}_k \ (l \le2)$. 

The last issue is the rank calculation. Denote a $(6+12m+6n)\times(15+4m+3n)$ observability matrix as $\mathcal{O}_{f} = \begin{bmatrix}
    \bar{\mathcal{O}}_{E_1}^T & \dots & \bar{\mathcal{O}}_{E_m}^T &\bar{\mathcal{O}}_{S_1}^T & \dots & \bar{\mathcal{O}}_{S_n}^T & \bigtriangledown \bm{b}_g^T & \bigtriangledown \bm{b}_a^T
\end{bmatrix}^T$, where $\bar{\mathcal{O}}_{S_k}$ and $\bar{\mathcal{O}}_{E_k}$ are defined in \eqref{basis}. The right null space of $\mathcal{O}_{f}$ is the unobervable distribution. Define the left null vector of $\mathcal{O}_{f}$  as
\begin{equation}
    \begin{bmatrix}
        \mathcal{LN}_{E^1} &\dots& \mathcal{LN}_{E^m}&\mathcal{LN}_{S^1} & \dots&\mathcal{LN}_{S^n} & \bm{0}_3^T & \bm{0}_3^T
    \end{bmatrix}
\end{equation} where $\mathcal{LN}_{E^k} = [\mathcal{LN}_{E^k,1},\ \mathcal{LN}_{E^k,2}, \ \mathcal{LN}_{E^k,3}, \ \mathcal{LN}_{E^k,4}]$ and $\mathcal{LN}_{S^k} = [\mathcal{LN}_{S^k,1},\ \mathcal{LN}_{S^k,2}, \ \mathcal{LN}_{S^k,3}, \ \mathcal{LN}_{S^k,4}]$. $\mathcal{LN}_{E^k,i}^T ,\mathcal{LN}_{S^k,1}^T \in R^{3}$ and $\mathcal{LN}_{S^k,j} \in R \ (j\ne 1)$.

Using Einstein summation convention, the equations are

\vspace{-4pt}

\begin{equation}
\resizebox{\columnwidth}{!}{$
    \left\{\begin{matrix}
  \mathcal{LN}_{S^k,2} =0 \ \ \ (\romannumeral1)
  \\
  \mathcal{LN}_{S^k,3}\bm{n}_k^T + \mathcal{LN}_{E^r,3}\bm{C}[\bm{l}_{r} \times] = 0 \ \ \ (\romannumeral2)
  \\
  \mathcal{LN}_{E^k,2}\bm{C}\bm{H}_{l_k} = 0 \ (\forall k)\ \ (\romannumeral3)
  \\
  \resizebox{\columnwidth}{!}{$
  \mathcal{LN}_{S^k,1} = k_{S^k,1}\bm{n}_k^T\bm{C}^T - \mathcal{LN}_{S^k,3}\bm{v}^T\bm{C}^T -  \mathcal{LN}_{S^k,4}\bm{g}^T\bm{C}^T \ (\forall k) \ \ (\romannumeral4)
  $}
  \\
\resizebox{\columnwidth}{!}{$
  \mathcal{LN}_{E^k,1} = k_{E^k,1}\bm{l}_k^T\bm{C}^T - \mathcal{LN}_{E^k,2}[\bm{C}(\bm{\Lambda}_{kx}\bm{\gamma}_{k1} +\bm{\Lambda}_{ky}\bm{\gamma}_{k2}+ \bm{p}) \times]+  \mathcal{LN}_{E^k,3}[\bm{Cv} \times] +  \mathcal{LN}_{E^k,4}[\bm{Cg} \times] \ (\forall k) \ \ (\romannumeral5)
  $}
  \\
  \resizebox{\columnwidth}{!}{$
  \mathcal{LN}_{S^k,1} [\bm{Cn}_k \times] + \mathcal{LN}_{E^r,1} [\bm{Cl}_r \times] + \mathcal{LN}_{E^r,2} [\bm{d}_{E_r} \times] + \mathcal{LN}_{E^r,3}[\bm{C}[\bm{l}_{r}\times]\bm{v} \times] +  \mathcal{LN}_{E^r,4}[\bm{C}[\bm{l}_{r}\times]\bm{g} \times]=\bm{0} \ \ \ (\romannumeral6)
          $}
\end{matrix}\right.
$}
\end{equation} Combining equation $(\romannumeral3)-(\romannumeral6)$ by Jacobi identity, 
\begin{equation}
    \begin{matrix}
\resizebox{\columnwidth}{!}{$
\bm{0}  = \mathcal{LN}_{S^k,4}\bm{g}^T\bm{C}^T[\bm{Cn}_k \times] + \mathcal{LN}_{E^r,4}[\bm{Cl}_r \times][\bm{Cg} \times] + \mathcal{LN}_{E^r,2}[\bm{Cl}_r \times][\bm{C}(\bm{\Lambda}_{rx}\bm{\gamma}_{r1} +\bm{\Lambda}_{ry}\bm{\gamma}_{r2}+ \bm{p})  \times] + (\mathcal{LN}_{S^k,3}\bm{n}_k^T\bm{C}^T+ \mathcal{LN}_{E^r,3}[\bm{Cl}_r \times])[\bm{Cv} \times]
  $}
        \\
        = -\mathcal{LN}_{S^k,4}\bm{n}_k^T\bm{C}^T[\bm{Cg} \times] + \mathcal{LN}_{E^r,4}[\bm{Cl}_r \times][\bm{Cg} \times] \ \ \ (\romannumeral6 ')
    \end{matrix}
\end{equation} the second equation is based on equation $(\romannumeral1)-(\romannumeral3)$.

When r(E) = 0, $(\romannumeral2)$ has $-r(S) + n$ additional degrees of freedom, $(\romannumeral4)$ has $n$ additional degrees of freedom and $(\romannumeral6 ')$ has $-r(S) + n + 1_{D_1}$ additional degrees of freedom. Hence, dim$(\triangle)$ = $(15+3n) - (6+6n - 3n - 1_{D_1}+ 2r(S)) = 9-2r(S)+1_{D_1}$.

When r(E) = 1, the degrees of freedom are as follows: $(\romannumeral2)$: $1_{D_3} + n + 3m - 3$, $(\romannumeral3)$: $m$, $(\romannumeral4)$: $n$,  $(\romannumeral5)$: $m$ and $(\romannumeral6 ')$: $n + 3m -2+ 1_{D_3}1_{D_2}$. dim($\triangle$) $=(15+6m+3n) - (6+5+12m+6n -8m- 3n - 1_{D_3} - 1_{D_3}1_{D_2}) = 4 + 1_{D_3} + 1_{D_3}1_{D_2}$. When r(S) = 0, $1_{D_3} = 1$.

When r(E) $\ge$ 2, the degrees of freedom are as follows: $(\romannumeral2)$: $ n + 3m - 3$, $(\romannumeral3)$: $m$, $(\romannumeral4)$: $n$,  $(\romannumeral5)$: $m$ and $(\romannumeral6 ')$: $n + 3m -2$. dim($\triangle$) $=(15+6m+3n) - (6+5+12m+6n -8m- 3n) = 4$.

\end{document}